\documentclass[twoside,11pt]{article}

\usepackage{jagi}
\usepackage{paralist}

\ShortHeadings{Measuring Intelligence through Games}{Schaul, Togelius and Schmidhuber}
\firstpageno{1}

\begin{document}

\title{Measuring Intelligence through Games}

\author{\name Tom Schaul \email tom@idsia.ch \\
            \addr IDSIA, University of Lugano\\
            Galleria 2\\
            6927 Manno-Lugano, Switzerland
            \AND            
       \name Julian Togelius \email julian@togelius.com \\
       \addr IT University of Copenhagen\\
       Rued Langgaards Vej 7\\
       2300 Copenhagen S, Denmark
       \AND
       \name J\"{u}rgen Schmidhuber \email juergen@idsia.ch \\
      \addr IDSIA, University of Lugano\\
            Galleria 2\\
            6927 Manno-Lugano, Switzerland
            }

\editor{N/A}

\maketitle

\begin{abstract}
Artificial general intelligence (AGI) refers to research aimed at tackling the full problem of artificial intelligence, that is, create truly intelligent agents. This sets it apart from most AI research which aims at solving relatively narrow domains, such as character recognition, motion planning, or increasing player satisfaction in games. But how do we know when an agent is truly intelligent? A common point of reference in the AGI community is Legg and Hutter's formal definition of universal intelligence, which has the appeal of simplicity and generality but is unfortunately incomputable.

Games of various kinds are commonly used as benchmarks for ``narrow'' AI research, as they are considered to have many important properties. We argue that many of these properties carry over to the testing of general intelligence as well. We then sketch how such testing could practically be carried out. The central part of this sketch is an extension of universal intelligence to deal with finite time, and the use of sampling of the space of games expressed in a suitably biased game description language.
\end{abstract}


\section{Introduction}
Artificial General Intelligence (AGI) is the bold endeavor of reaching beyond
the narrow focus of much of the current artificial intelligence research community,
aiming to build agents that encompass the whole breadth of human intellectual faculties and more.
In contrast to narrow AI, the focus of AGI research lies on the breadth of the range of environments in which an agent behaves intelligently,
rather than the performance in a particular environment. Ultimately, that range should cover all environments that
humans can act in, and more.

While artificial intelligence originally had this broad vision (\cite{Schmidhuber:06ai75}), many practitioners have taken to
specializing on particular, more manageable subproblems. While this trend has greatly advanced the field
and been highly successful for many practical applications, many find it philosophically unsatisfactory.
Not surprisingly therefore, the pendulum has started to swing back, and there is a resurgent interest
in the big questions on how artificial \emph{general} intelligence can come about.
The recent development of a solid theoretical framework for AGI by \cite{hutter05universal}
has played a major role in this rekindling.

Alongside this development there has been an increased effort toward designing an objective and practical benchmark
for measuring general intelligence, because it would allow for a better comparability between the very diverse approaches to AGI
and homogenize the field.
While the debate is slowly converging around the recently proposed formal measure of \emph{universal intelligence} by \cite{Legg2007},
no truly general yet practical benchmark has been established.

A general benchmark will necessarily need to evaluate an agent on a large set of environments,
and in order to form a single composite score, each environment must have an associated weight.
In the best case, the set of environments includes \emph{all} well-defined, practically evaluable environments.
In practice, however, we will have to restrict this set, thereby introducing a bias and making the benchmark less general.
Concretely, this paper proposes a benchmark limited to the subset of \emph{game} environments.
We will argue for this particular choice, showing that it preserves a high level of generality,
while at the same time being practically useful.

We will start by introducing measures of general intelligence, how they can be altered to include
resource constraints and how they implicitly determine a weighting of the set of environments (section~\ref{intelligence}).
We then discuss the suitability of games as a class of environments (section~\ref{gamebench}),
before connecting the dots and defining a benchmark for measuring general game intelligence (section~\ref{gameintel}).
Finally, we tackle the issue of game description languages, and how existing ones could be used (section~\ref{gamelang}).

\section{Defining Intelligence}
\label{intelligence}
Intelligence is one of those interesting concepts that everyone has an opinion about, but few people are able to give a definition for
-- and when they do, their definitions tend to disagree with each other.
And curiously, the consensus opinions change over time: consider for example a number of
indicators for human intelligence like arithmetic skills, memory capacity, chess playing, theorem proving --
all of which were commonly employed in the past, but since machines now outperform humans on those tasks, they have fallen into disuse.
We refer the interested reader to a comprehensive treatment of the subject matter in~\cite{Legg2008}.

The current artificial intelligence literature features a panoply of benchmarks, many of which, unfortunately, are very narrow,
applicable only on a small class of tasks. This is not to say that they cannot be useful for advancing the field,
but in retrospect it often becomes clear how little an advance on a narrow task contributed to the
general field. For example, researchers used to argue that serious progress on a game as complex as chess
would necessarily generate many insights, and the techniques employed in the solution would be useful for real-world problems -- well, no.

All this highlights the need for a very general definition that goes beyond an aggregation of a handful of tasks.
We now introduce the most general definition to date (section~\ref{univintel}), which unfortunately is only of theoretical use,
as the quantities it relies on are not computable. In section~\ref{time} then, we delineate a more practical version that takes computation time into account.

\subsection{Universal Intelligence}
\label{univintel}
Building upon Solomonoff's theory of universal induction~\cite{Solomonoff:64},
and extending it to handle agents that act in their environment (in contrast to just passively pondering upon observations),
\cite{hutter05universal} recently developed a formal framework for universal artificial intelligence.
Within the very general reinforcement learning setting, he formally describes an optimal
agent (called AIXI) that maximizes the expected reward for all possible unknown environments --
the only caveat being its incomputability.

As a dual of this framework, \cite{Legg2007} define a formal measure
of universal intelligence, for which AIXI, per definition, achieves the highest score.
Interestingly, their resulting definition coincides well with many informal, mainstream ones.
Here we summarize their results, as they will form the theoretical basis for much of this paper.

\begin{verse}
\emph{Intelligence measures an agent's ability to achieve goals in a wide range of environments.}
\end{verse}

Formally, the intelligence measure $\Upsilon$ of an agent $\pi$ in a class of (computable) environments $E$ is defined as
\[
\Upsilon(\pi) := \sum_{\mu \in E} 2^{-K(\mu)}V_{\mu}(\pi)
\]
where $V_{\mu}({\pi})$ is the expected total reward of $\pi$ when acting in a particular environment $\mu$,
and $K(\mu)$ is a complexity measure of the environment (satisfying the technical condition  
$\sum_{\mu \in E} 2^{-K(\mu)} < \infty$).
We call $\Upsilon(\pi)$ the \emph{universal intelligence} of $\pi$ when $E$ is the set of all computable environments,
and $K(\mu)$ is the Kolmogorov complexity of $\mu$, i.e. the
length of the shortest program that fully describes $\mu$.

Very informally, this equation can be read as ``the universal intelligence of an agent is the sum of how well it performs in all computable environments, logarithmically weighted by their complexity so that simpler environments count more''.
For a more intuitive understanding, it may be useful to expand upon a number of aspects of this definition:
\begin{itemize}
\item {\bf Environments}: We take the term environment to encompass not only the dynamics that define what happens for each possible action
(which might in turn include the reaction of an adversary),
but also the rules according to which the agent is rewarded.

\item {\bf Goals}: According to the most well-received theory of the development of life on earth (Darwin's),
all living beings can be said to have a goal (maximization of evolutionary fitness), but as this is not necessarily true, i.e. not true in all possible worlds,
it is debatable whether goals are an existential requirement for intelligence.
While we do not claim that intelligence in the broad sense cannot in part be purely `contemplative',
we cannot conceive of how the intelligence of an agent that is devoid of any goal-driven behavior could be measured.
For practicality therefore, we assume that each environment produces a numeric `reward' value,
and maximizing it is the goal of an intelligent agent.

\item {\bf Acting in the environment}: In order for intelligence to be measurable, an agent must act,
as only actions can be evaluated objectively, not its internal processes (like awareness).

\item {\bf Space of environments}: The presented definition can be seen as an intelligence measure,
also if the set $E$ is more restrictive, e.g.~limited to a
particular domain of interest, tasks where humans excel, say.
This is related to the notion of pragmatic general intelligence by \cite{Goertzel2010}.

\item {\bf Weighting by complexity}: We need some way to assign relative importance to different environments,
which is done through a complexity measure $K$, traditionally measured in bits\footnote{Note that there is no way to avoid non-uniform weighting: There exists no uniform probability distribution on the integers.}.
All environments are computable, therefore can be concisely represented as a shortest piece of code -- the length of this code is the environment complexity.
\end{itemize}

Legg and Hutter propose their definition as a basis for any test of artificial general intelligence.
Among the advantages they list are its wide range of applicability (from random to super-human),
its objectivity, its universality, and the fact that it is formally defined.

Unfortunately however, it suffers from two major limitations:
\begin{inparaenum}[\itshape a\upshape)]
\item {\em Incomputability}: Universal intelligence is incomputable, because the Kolmogorov
complexity is incomputable for any environment (due to the halting problem).
\item {\em Unlimited resources}: The authors deliberately do not include any consideration
of time or space resources in their definition. This means that two agents that act identically in theory
will be assigned the exact same intelligence $\Upsilon$, even if one of them requires infinitely more computational
resources to choose its action (i.e. would never get to do any action in practice) than the other.
\end{inparaenum}

There have been a number of attempts to overcome the limitations of AIXI.
\begin{enumerate}
	\item We may replace Solomonoff's incomputable prior
by the {\em Speed Prior} (\cite{Schmidhuber:02colt}),
which assigns high probability to quickly computable
environments (instead of those with the shortest
descriptions / lowest Kolmogorov complexity favored by
Solomonoff's prior).
This yields a computable agent AIS which can predict
expected reward with arbitrary accuracy in finite time.
	\item We may use a G\"{o}del Machine with
a fixed limit of computational resources per action
(\cite{Schmidhuber:03gm,Schmidhuber:09gm}).
	\item We may use a Monte-Carlo approximation of AIXI (\cite{Veness2009}).
This already yielded promising practical results on an ad-hoc portfolio
of simple maze-tasks and games (including Tic-Tac-Toe and Pac-Man):
the same AIXI-approximating agent learned to act reasonably well
in all of them.
\end{enumerate}

While it is clearly a useful direction to derive practical, scaled-down variants of uncomputable,
universally optimal agents, here
we are concerned with the dual case: Rendering the \emph{definition} practical.
One way of doing this would be to rephrase the definitions of Legg and Hutter
in the context of the Speed Prior (\cite{Schmidhuber:02colt}) instead of
Solomonoff's prior.
In the next section, however, we will follow an even more pragmatic approach:
we will greatly limit the class of environments further such that each member of
the class is not only quickly computable, but also of obvious
interest to a wide community of people, namely, gamers.

\subsection{Intelligence and Limited Time}
\label{time}

Clearly, any practically useful measure of general intelligence (i.e.~that yields a good result in finite time)
needs to take computation time of the agent and environment into account.
Also, we posit that all agents are computable, and all environments are episodic (i.e.~finish after a finite number of interactions).
The episodic assumption should be rather uncontroversial, as all known organisms have finite life spans, and it is unclear what it would mean to behave well in a task that might never finish. The computability assumption is necessary for anything running on standard computers.

Although other resources, like memory, may be of practical importance as well, in the following we limit ourselves to only time.
(Most of the arguments remain applicable to other limited resources.)

The two aspects of computation time to consider are:
\begin{enumerate}
\item The time the \emph{environment} requires to generate the next state/observation, after each action.
We propose to incorporate the expected time into the environment's complexity measure.
A standard formulation for this would be the equivalent of Levin complexity:
\[
K(\mu) = l(\mu) + \log(\tau(\mu)),
\]
where $l(\mu)$ is the (not necessarily shortest) length of the description of the environment $\mu$ and $\tau(\mu)$ is the expected computation time of $\mu$ (for one episode)\footnote{Usually, and for asynchronous environments in particular, $\tau(\mu)$ is an unproblematic value to define, it could even be set to the total time budget $T$, see below. In some cases however, the computation time of the environment may depend on the agent's actions (e.g. updating a physical simulation when an agent jumps into a liquid should take longer to compute than if the agent stands still). We sidestep this issue by taking $\tau(\mu)$ with respect to a randomly acting agent. This may be reasonable however, because as the inherent complexity of the environment increases, the actions of the agent arguably account for a relatively smaller variation in $\tau(\mu)$ (which in turn contributes relatively less to $K(\mu)$), thus becoming irrelevant in the limit.
}.
This corresponds to a trade-off that weights length exponentially more heavily than time.
Many other trade-offs are possible (see e.g.~\cite{sun2010frontier} for an extensive discussion).

\item The time the \emph{agent} requires to decide upon an action.
We propose to incorporate this into the definition of intelligence itself.
\end{enumerate}

There are two ways for integrating resource limitations into a general intelligence measure:
Either the agent has an unlimited budget of time, but its performance is somehow weighted
by the resources it consumes (see e.g.~\cite{Goertzel2010} for a simple variant of this),
or the agent has a limited budget, and the intelligence measure is based on the (cumulative or average)
reward it can achieve within that limit (e.g.~\cite{Hernandez-Orallo2010,Schmidhuber:09gm}).
The first of these is seriously flawed, as it can be exploited by `hyperactive' agents, that act extremely fast and randomly,
with the effect of multiplying the low reward by a large number.

In addition, resource consumption can be handled through the reward function,
e.g. if there is an implicit assumption that if completing an episode faster is valued more,
then this can be reflected in the rewards directly.

\subsection{Two-phase Evaluation}
Given an environment $\mu$ and a total time budget $T$\footnote{As far as possible, the time units should be invariant with respect to hardware of implementation details, e.g. it could be the number of CPU instructions executed.} for the agent, we propose to set up the measurement as follows.
The time is split into a \emph{learning phase} and an \emph{evaluation phase}, and the agent chooses itself when to (irreversibly) switch to the latter.
Rewards gained during the learning phase are disregarded, and the environment is reset at that point
(otherwise an agent could exploit by switching just before a predicted success, and stop afterward).
The final reward $V_{\mu, T}(\mu)$ is the average reward of the completed episodes
during the evaluation phase (we need to take the average instead of the sum, here again, to handle `hyperactive' agents),
the accumulated reward so far if no episode was completed,
or zero if the agent never switched.

The motivation for having two phases is the issue of \emph{learning}.
Given that a task is potentially attempted more than once,
it is debatable whether the average performance is what defines intelligence,
or instead the capability to improve between early trials and later ones.
Applying the latter naively is prone to exploitation: That agent would be judged most intelligent which can best hide its true skill in the first episodes
(deliberately acting as bad as possible) while acting normally in the end.
Requiring the agent to switch itself removes this moral hazard.
On the other hand, using average performance over all attempts is
a relatively harsh setting for the exploration-exploitation trade-off,
which may force the agent to act overly carefully, not learning enough about the environment,
and therefore leading to an underestimation of its intelligence,
a problem alleviated by the learning phase.
Importantly, this setup allows us to compare the intelligence of very different types of agents, all on the same scale:
a reasoning-based agent (that does not learn) can use all available resources for that, skipping the learning phase,
while an evolution-based agent could employ most of those resources to evolve a good policy during many quick episodes and have itself be evaluated
on the best one encountered. In between those extremes, a good reinforcement learning agent could handle the exploration-exploitation trade-off explicitly, including the switching action.

Interactions could be synchronous (i.e.~the environment is paused until the agent chooses an action), but should preferably be asynchronous (`pass' actions until the agent has made a decision), as it entails a more natural and environment-specific penalisation of slowly acting agents.

\subsection{Anytime-measure}
In the best case, a practical intelligence measure should be an `anytime'-measure,
i.e.~that can be stopped anytime, and gives more accurate results the longer it runs.
A simple but effective way to achieve this is to use a Monte-Carlo estimate of $\Upsilon$,
sampling more and more environments (that form a set $\hat{E}$), where the probability of $\mu$ being sampled is proportional to $2^{-K(\mu)}$.
Like most Monte-Carlo-based methods, this intrinsically easy to parallelize.
\[
\widehat{\Upsilon}(\pi) := \frac{1}{|\hat{E}|} \sum_{\mu \in \hat{E} \subset E}
V_{\mu, T}(\pi)
\]

Note that this assumes that the time limit $T$ is specified within the environment description itself,
and thereby incorporated in the task complexity
(larger total budgets corresponding to more complex environments).
If $T$ is not part of the complexity, the following iterative scheme can be shown to be an equivalent alternative.
Start with $T=T_0$, $i=0$. In each iteration $i$, evaluate $2^i$ environments, half of which for $T=2^{0}T_0$, a quarter for $T=2^{1}T_0$, and so forth, and one for $T=2^{i}T_0$.

To summarize, we adapted the definition of universal intelligence in three places to make it practically useful:
\begin{enumerate}
\item We replaced Kolmogorov complexity with a computable complexity measure that penalizes heavy computational requirements for the environments,
\item we incorporated resource usage into the performance measure of an agent to favor efficient ones while avoiding some well-known pitfalls,
\item we formalized a Monte-Carlo approximation that can handle the infinite set of environments, while at the same time forming an anytime-measure.
\end{enumerate}

The next sections will add the final missing piece, namely a suitably biased domain of environments, which are at once meaningful, easy to evaluate and easy to sample from: Games.

\section{Games as Testbeds}
\label{gamebench}

Games and artificial intelligence have close and long-standing ties to each other. Turing himself,
who proposed the first test for machine intelligence, also invented the MiniMax algorithm for perfect
information two-player games, and considered chess an important domain for future computer science
research (\cite{turing50computing}). What is arguably the world's first reinforcement learning algorithm (a precursor
to modern temporal difference learning) was invented by Samuel in the context of building an automatic
checkers playing program, thus kickstarting a major strand of modern AI research.

More recently, several high-profile AI researchers have proposed games as good benchmarks for
AI (\cite{laird01human,Masum2003}). At least part of the argument is that the technical development
of modern computer games has now definitely overtaken custom-built benchmarks and robot simulators,
and the commercial game industry provides a huge amount of high-quality, well-tuned problems and
environments for AI research as a side effect of the commercial pressure for better games.

\subsection{What are games?}

There is not one definition of what a game is, but plenty. In fact, Wittgenstein used the concept of a game in a number of thought-experiments designed to show that it was impossible to correctly define any concept in terms of sufficient and necessary conditions; instead, concepts are implicitly defined by those things that they refer to, and which are related to each other through family likeness. Learning to use a concept is learning to play the language-game that the concept forms part of, yet another example of a game (\cite{wittgenstein53philosophical}).

The impossibility of defining a naturally occurring concept does not mean you should not try; a working definition of games would be great, even if we acknowledge that it's not all-encompassing. Game designer Sid Meier defines a game as ``a series of meaningful choices''. Others, such as \cite{salen04rules}, emphasize conflicts as central to games: A game is ``a system in which players engage in an artificial conflict, defined by rules, that results in a quantifiable outcome''. \cite{juul05halfreal} provides a more formal definition: ``A game is a rule-based formal system with a variable and quantifiable outcome, where different outcomes are assigned different values, the player exerts effort in order to influence the outcome, the player feels attached to the outcome, and the consequences of the activity are optional and negotiable''.

The above definitions are not without their critics. For example, game designer Raph Koster remarks that none of them contain the word ``fun'' (\cite{koster05a}). As fun seems to be so central to games, he then devotes a whole book to understanding what makes games fun.

For our purposes, a more relevant criticism of the above definitions is that they refer to a ``player'' who can ``exert effort'' and ``engage'' in ``meaningful choices''. Obviously, including such a player in a definition of a game which is used to test artificial intelligence would beg the question of artificial intelligence as a whole. We therefore choose to adapt Juul's definition to fit our purposes:
\begin{verse}
    \emph{A game is a rule-based formal system with a variable and quantifiable outcome, where different outcomes are assigned different values, and an agent can affect the outcome through taking actions during the game based on full or partial knowledge of the game state.}
\end{verse}

\subsection{Types of Games}

There are countless types of games, and more different taxonomies of games than we can do justice to here. Even if we restrict ourselves to games that can be played on a computer (which excludes football, but not football simulations) there are numerous genres, which are sometimes so different that it's remarkable that the same word is used to all of them: card games, board games, mathematical games, first-person shooter games, real-time strategy games, flight simulator games, quiz games, role-playing games, puzzle games, rhythm games, virtual world games and so on. Instead of attempting another taxonomy, we can draw up a few dimensions along which games can vary, together with short comments on how the requirements on the agent vary with these dimensions:

\begin{itemize}

\item {\bf Observability}: Perfect information games like chess allow the agent to see the whole game state at any time, whereas a game with very low observability like Battleships starts with the agent in complete darkness and drip-feeds of information about the game state. A perfect information game can be played by a reactive agent, whereas a game with low observability requires the agent to make hypotheses about the game state and remember past information.

\item {\bf Input dimensionality and representation}: The agent's knowledge of the state space in Poker can be represented as a short sequence of integers representing the cards at hand, whereas the high-resolution 3-dimensional image which constitutes an observation in Halo requires several megabytes of data to represent accurately. The latter representation requires the agent to perform sophisticated processing of visual input, something which some researchers consider an integral part of intelligence (e.g. many researchers in the adaptive behaviour community claim that it's meaningless to study intelligence except when grounded in a complete sensorymotor loop) and other researchers consider a peripheral distraction.

\item {\bf Single-, duo- or multiplayer}: Single-player games are those where you only play against the game itself, but many games have two or more players; some have many millions of players (e.g. Farmville or World of Warcraft). The extent to which the players compete directly against each other varies from those games where only one player can win, to those where players mostly collaborate. The presence of other players in the game will usually require the agent to model and predict the other players' actions in order to do well in the game.

\end{itemize}

\subsection{Arguments for Games as AI Testbeds}
\label{gameargs}

So why would games be better as testbeds for AI than other problems, say, theorem proving, image recognition, robot navigation or natural language interaction? A large number of arguments have been put forward, some of which only apply to some types of games and some aspects of intelligence, others which are more general.

\begin{itemize}
\item {\it  Simplicity}: Many games can be concisely described using unambiguous mathematical notation, and have surprising depth despite their apparently bare-bone dynamics. A good example of this is the ancient board game Go, which offers enormous depth despite the rules being so simple they can be written in a few lines.
\item {\it  Natural reward function}: Games are typically constructed so that they have a very natural reward function: the score. Score-based reward functions are often smooth and fine-grained, as many different actions and events are scored. This means that the player's proficiency can be accurately quantitatively measured for both high-skilled and low-skilled players.
\item {\it  Scalability}: Most games are made to be played by humans, and to be learned by humans while playing. This means that they typically possess a human-friendly long and smooth learning curve, ideal for reinforcement learning. Agents of very low skill can make some progress (better than agents of no skill) while agents of higher skill can reach much higher performance.
\item {\it  Proven performance and remaining challenge}: Games is a domain where on some instances (chess, checkers, backgammon, etc.) computers have already reached or surpassed human-level performance, suggesting that AI approaches can learn to beat humans at other games. At the same time, there remain considerable challenges, as computer programs are nowhere near competitive with the best players of e.g. Go or StarCraft.
\item {\it Understandability}:  Playing games is an activity humans understand, making it easy to understand what the agent is doing and how well it is doing it.
We can also judge the performance of an agent by playing directly against it.
 \item {\it  Fun}: People like games, both playing, watching and talking about them.
This goes for researchers, students and ordinary people.
Therefore, it's often easier to find students for game-related research than for other topics.
\item {\it  Public awareness}: Most people know what games are about, unlike more abstract tasks (like theorem-proving), which makes it easier to explain what a research breakthrough is about.
\item {\it  Industrial applicability}: There's money in games, as the game industry is already the largest of the entertainment industries, and keeps growing.
Games and virtual worlds are more and more used for training, planning and education purposes.
Therefore, it can be easier to get funding for game-related research than for research on AI applications in other domains.
\item {\it Availability and cost}: Game developers and hobbyists have made implementations of all types of games widely available, either for free or for a very modest sum. Some games come with extensive interfaces for interfacing AI in various roles and for modifying the games in various ways (e.g. Civilization IV and Unreal Tournament 2004). For some older games (e.g. Quake) and many hobbyist-developed games the source code is freely available.
\item {\it Speed}: Most games are relatively quick to play even when played in ``real-time''. Many video games can be sped up to be played thousands of times faster than real-time on current hardware; e.g. the current Mario AI Competition uses a version of Super Mario Bros where a couple of hundred games can be played per second. Board games and mathematical games have no ``real-time'', and the simplicity of evaluating the game mechanics means that millions of games can be played per second and virtually the only limiting factor is the computational complexity of the player.
\item {\it  Diversity}: As noted above, there are innumerable genres of games; arguably, there exist games related to virtually every human cognitive task.
For example, many games require cognitive skills such as visual and auditory perception, communication, cooperation and competition,
planning and reasoning, navigation and mapping, or prediction and model-building.
\end{itemize}

Note that the most commonly touted classes of benchmark problems/environments fall far short of games on several of the dimensions outline above. For example:

\begin{itemize}
\item Physical robotics: simplicity, natural reward function, scalability, proven performance, availability and cost, speed, diversity.
\item Simulated robotics:  simplicity, natural reward function, scalability, proven performance, public awareness, diversity.
\item Theorem proving: scalability, proven performance, understandability, fun, public awareness, industrial applicability, diversity.
\item Natural language understanding/production: simplicity,  natural reward function, understandability, fun, public awareness, availability and cost (of data),  diversity.
\item The original Turing test: scalability, proven performance, availability and cost (of test participants), speed, diversity.
\end{itemize}


\subsection{Competitions}
\label{compintro}

Game challenges for artificial intelligence are often posed in the form of competitions. The world's probably most famous AI event was a games-based competition: the match between former chess world champion Garry Kasparov and the IBM Deep Blue integrated software/hardware. The victory of machine over flesh in this competition prompted an uncommon but vigorous and certainly welcome public discussion about the nature of intelligence and whether if could be embedded in a machine. A near-consensus among commentators was that because this seemingly very complicated problem (playing chess) could be solved with just a simple search algorithm and a massive database, it was clearly not necessary to be intelligent in order to be able to play chess. (This is an example of the phenomenon of the ever-moving goalpost for AI: as soon as AI techniques are shown to solve a problem, the problem is deemed not requiring intelligence, and its solution becomes ``mere computer science''.)

Since then, game AI competitions have diversified significantly. A number of games-based AI competitions are currently running and enjoying healthy
numbers of submissions from academic AI researchers with various specialties.
In particular, the IEEE computational intelligence society sponsors a number of competitions
associated with its Congress on Evolutionary Computation and Conference on
Computational Intelligence and Games. These competitions are based on submitting
agents that play particular games well; the very diverse collection of games used
includes Othello, Go, Pac-Man, Super Mario Bros, Unreal Tournament and
a car racing game (\cite{loiacono08the,loiacono10the,togelius10the,togelius08the,hingston09a}).
The point of these competitions is to focus researchers' efforts on a single problem,
which has not been crafted in order to favour a particular algorithm,
resulting in a reasonably fair and reliable comparison of competing AI algorithms.

One competition we will find reason to return to in more depth later on is the
Stanford General Game Playing Competition, which differs from the above competitions
in that agents are judged not on their ability to play a single game, but a number of unseen games.
These games are described in a Game Description Language, which will be discussed further in section~\ref{gamelang}.

\section{General Game Intelligence}
\label{gameintel}

Motivated by sections~\ref{gameargs} and~\ref{compintro},
we now proceed combine the practical measure of intelligence from section~\ref{time} with a space of environments that is restricted to games,
leading us to a practical measure of general game intelligence.
In particular, we put the following restrictions on the set of environments:
\begin{itemize}

\item The total number of interactions is guaranteed to be finite (all games end eventually).

\item The sum of rewards achievable in an episode (i.e.~the game-score) is bounded.

\item The agent-environment interface is simple: the environment sends a
string of symbols as observation, and the agent sends back a string of symbols as its action.

\item Each game is encoded in a game description language (see section~\ref{gamelang} for details),
and the length of this encoding, together with the computational resources required to run
the game define its complexity $K$ (as described in section~\ref{time}).
The assumption is that short encodings correspond to simpler games.

\item When the game allows for a fixed adversary (e.g.~Deep Blue),
the encoding length of the adversary, as well as its computation time, get incorporated in the complexity measure of the game,
as if they were part of the environment. This automatically makes games with stronger opponents more complex,
and adjusts their weight in the total intelligence measure accordingly.

\end{itemize}

We distinguish two classes of game environments: Those that interface to a single agent,
for which we then define an \emph{absolute} measure of general game intelligence,
and those with a higher number of players (typically 2) for which we similarly define a
\emph{relative} measure.
This relative measure of general game intelligence can then be used
to establish a ladder system or a unique ranking of all participating agents (e.g.~Elo). 
This may give a richer description of the capabilities of an agent than the single number generated by the absolute measure,
but is not quite as objective.

\section{Game Description Languages}
\label{gamelang}

A game description language (GDL) is a language in which games can be described.
More formally, each GDL is accompanied by an interpreter which transforms GDL strings into games,
and a valid GDL string is one which can be transformed into a game by this interpreter.

One could argue that any programming language constitutes a game description language,
as would a universal Turing machine. However, a game in the sense we consider it here
needs to have specified channels for input, output and reward signals, which is not true of programs in general.
If we arbitrarily assign these channels to e.g. parts of memory or positions on the program tape,
infinitesimally few of all valid programs would also be games in any meaningful sense.

Existing GDLs are much more limited in what they can express,
in that they are not Turing-complete, and cannot even express all possible games.
Even within the space of games which they can express, they are
biased towards particular types of games (sampling all valid strings of a
particular length will yield some types of games more often than others).
The following are some existing game design languages:

\begin{itemize}
\item The Stanford GDL is the language used in the Stanford General Game Playing Competition (\cite{Love2008}). This language is based on defining objects, actions and relations (representing legality of moves, effects etc.), and could in principle define a very large number of different games, though it is biased towards board games. It is limited to perfect information games with a discrete and finite state space, though it is claimed that it could be extended to imperfect information games. As a result of being so low-level, the Stanford GDL is not particularly compact, even when defining the type of games which it is biased towards; the example definition of Tic-Tac-Toe runs to two pages.

\item The Ludi language was invented by Browne and Maire and used in their work on automatically generating games using evolutionary algorithms (\cite{Browne2009}). This language is restricted to ``recombination games'' (essentially a subset of board games) and is structured into a tree form, similar to LISP. Each branch of the tree is a {\it Ludeme} which describes a particular aspect of the game, such as the shape of the board or number of units. Due to being more specialized it is often quite compact, e.g. Tic-Tac-Toe can be defined in six lines.

\item Another GDL with a more narrow domain is the language used by \cite{togelius08experiment} in their experiment on evolving Pac-Man-like games. This language only admits fixed-length strings, where each string position has a particular meaning, e.g. the number of entities (or ``things'') of a particular colour, the movement logic of some entity, and what happens when a particular type of entity collides with another. The language is limited to predator-prey-like games in a discrete space of a given size.

\end{itemize}

What, then, would be desirable properties of a GDL used for testing for artificial general intelligence? The following are some (potentially conflicting) suggestions:

\begin{itemize}
\item {\it Expressiveness}: The language should be able to express a large variety of different games,
 in order to test as many aspects of general intelligence as possible. It is desirable that the GDL should
be able to express finite state as well as continuous state games, large as well as small inputs spaces,
single-player games as well as multi-player games, perfect information games as well as partial information games,
deterministic games as well as those with noisy state transitions, etc.

\item {\it Compactness}: The representation of any particular game should be as short as possible.
This also entails a language that is easy to sample from: many random strings will be valid games.

\item {\it Meaningfulness}: As high a fraction as possible of possible games should neither be trivial nor impossible,
but have a significant skill differentiation in their outcome.

\end{itemize}

It is worth noting that there are likely to be partial conflicts between these properties, so that e.g. more compact games would likely be less expressive; however, clever design efforts will probably be able to find languages that satisfy all properties to a reasonable degree.

It is also worth noting that it is perfectly possible to devise languages which describe games with much more complex state spaces and input/output representations than the type of games described by current GDLs. There is no restriction on the complexity of the interpreter which generates games from descriptions; complete game engines, such as {\it Unity} or the {\it Unreal} engine could be included along with component artwork. This would allow the description of e.g. first-person shooter games and real-time strategy games.
 
\section{AGI and Game Competitions}

If intelligence tests along these lines are to be realized, one natural way to do it
(especially for determining relative measures of general intelligence) would be in the form of public competitions.
As discussed above, a number of competitions are currently ongoing (and many more have run in the past) where AI techniques are tested using games of various types. However, with the exception of the Stanford General Game Playing Competition, they are all measuring performance on a single benchmark only. The Stanford competition has met with rather limited participation, and is very different in setup from the ideas proposed here in several ways -- most importantly, the agents are given a complete description of the environment, favoring reasoning over learning approaches.

It is important to state that we are not advocating an immediate transition to a unique intelligence test and a single competition based on it.
Rather, the ideas in this paper could form the basis for a number of separate competitions, possibly based on already existing competitions.
These competitions would be based on different GDLs, that are more or less strongly biased towards particular types of games.

One way to achieve a smooth transition from a game-specific competition to a more general competition
could be to take an existing benchmark game, and break it down into a number of components and parameters,
which can be used to build a game description language. This GDL could then describe the original game,
subsets of the original game and variations of it. For example, one of the currently ongoing competitions is
about constructing an AI controller that plays Super Mario Bros\footnote{http://www.marioai.org, ~\cite{togelius10the}}.
This classic game could maybe be decomposed into ``ludemes'' describing e.g. what happens when
Mario runs into enemies of various types, rules determining behavior of the various NPCs,
effects that power-ups have, the physics of the game,
movement capabilities of Mario, sizes and topologies of levels, rules about scoring, winning and losing, etc.
This would allow the expression of a number of games being somewhat similar to Mario, but differing
greatly in game mechanics to be expressed in a language that could be sampled according to the ideas expressed in this paper.

\section{Objections}
In this section, we gather a few conceivable objections to the central proposition in this paper, along with our responses to these objections.

\begin{itemize}
\item {\it ``This test will have enormous computational requirements. Playing many episodes of so many different games is just not feasible.''}

Measuring the performance on all possible games is certainly impossible. But our test is a sample-based anytime measure, meaning that the more time we have, the more games we can test on and therefore the more accurate our measure becomes. If we have extremely limited time and/or computer power available we can test an agent on only a handful games. If we want to compare the intelligence of two agents in very limited time, however, it is important that we test them on the same games in order to counteract the effects of having chosen this particular subset of games. Limiting the evaluation time may however bias the test toward particular aspects of intelligence, e.g. when it is too short for learning to be effective.

\item {\it ``Most games drawn at random will be meaningless, and winning or losing them will not be indicative of any interesting form of intelligence.''}

This partly depends on the game description language chosen; some GDLs produce a higher proportion of languages with good skill differentiation than others. But even if a large majority of sampled games are trivial or impossible, or only test very narrow abilities, this will not bias the test significantly. All agents will fail on the impossible games, and all agents of minimal intelligence will learn to solve the trivial or narrow games. Therefore, the ranking of various agents will be decided by their performance on the more interesting games. At most, the results of the test may be renormalized w.r.t.~the proportion of trivial and impossible games, for readability. It might also be possible to automatically pre-test the games for basic playability and learnability before using them as part of the test.

\item {\it ``Is is not counter-intuitive to assign higher weight to simple environments,
which presumably require less intelligence to tackle?''}
The reason behind this is two-fold. First, there are many more complex games than simple ones,
so taken together, the weight of the complex ones is substantial. Second, this helps avoid over-specialization,
as each general agent is required to handle most simple environments -- also intuitively,
we can question the \emph{general} intelligence of the clich\`{e} math genius who is unable to tie his shoes. And as stated above, if any game is so easy that all agents can play it, the difference between the agents will be decided by the more difficult games.

Note that a short description does not necessarily imply a simple environment:
complexity can emerge from simple rules (e.g. fractals, the game of Go).

\item {\it ``Your test will only measure very narrow aspects of intelligence, e.g. combinatorial reasoning. You completely miss out on the most important aspects of intelligence, namely (insert property here).''}

There is some substance to this objection, depending on the game description language chosen.
If the GDL is only capable of describing games within a rather narrow domain, the skills needed to
win at games within this domain will of course be more thoroughly tested. However, note that even games
with very simple mechanics appear to require sophisticated skills; cf. the difficulties faced by
AI researchers in building bots that outperform humans in both Go and Poker.
Bear in mind  that these seem to be very different problems requiring different approaches: An agent that could
play both games (which could certainly be represented by the same GDL) at high level would arguably
be more generally intelligent than anything we have now.

For more on how about this sort of test measures supposedly human capacities like intuition and creativity, see the extensive discussion in the original paper on universal intelligence (\cite{Legg2007}); the basic argument is that to the extent these capabilities help the agent solve the problem, they are implicitly tested.

\item {\it ``Your test is completely disembodied. But intelligence is a property of an embodied agent situated in an environment, grounding its symbols in direct interaction with the environment through sensors and actuators.''}

Whether embodiment is necessary is a debated topic, with differing views among researchers and philosophers from different camps. If we restrict the GDL to combinatorial or similar games, we are certainly leaving embodiment out. But to the extent that an agent can be said to be embodied in a virtual world, embodiment can be accommodated within this framework by simply including a 3D game engine in the GDL interpreter. This would make it possible to describe games taking place in rich 3D environments, forcing the agent to interpret high-dimensional visual input and map its own movements between body-space and world-space.

\item {\it ``You're not saying anything new. This is all implicit in the Stanford General Game Playing Competition.''}

There is, somewhat surprisingly, very little theoretical justification to be found for the Stanford GGP, at least in published form. In this paper, we lay out a theoretical framework for that competition and similar competitions, and connect it to a well-known theoretical contribution from the AGI community. We also discuss the severe limitations of the GDL used in that competition, and propose a principled way of sampling which games to play. Finally, a crucial difference between the Stanford GGP and the test we are proposing is that the Stanford GGP provides agents with the GDL specification of the game they are playing (arguably making the competition more about parsing and internal simulation than learning based on experience) whereas we propose to let the agents explore the game by interacting with it, which is more general.

\item {\it ``You're not saying anything new. This is all implicit in Legg and Hutter's definition of universal intelligence.''}

Universal intelligence is incomputable and does not take finite resources into account; it is expressly meant to be approximated, as a basis for more practical tests (a call which this paper is answering). The handling of finite resources through the agent managing a time budget and deciding when to switch between training and evaluation phases is new, as far as we know, as is the idea to sample the space of a GDL based on description length.

\end{itemize}

\section{Conclusions}

This paper we discussed why games should -- and how they can -- be used for research in Artificial General Intelligence.
We note that games are in many ways ideal for AI research,
but that current research which focuses on testing algorithms
on particular games fails to test for general intelligence: A general AI agent instead needs to be tested
on an appropriate, broad selection of unseen games.

To this effect, we have derived a practical measure of general game intelligence
from the Legg-Hutter definition of universal intelligence
which elegantly incorporates the usage of computational resources of both the agent and the game engine,
in addition of being an easily approximable anytime-measure.
The central idea then is to use length- and
resource-weighted sampling of strings from a game description language and evaluate the agent on the corresponding games.
As game description languages are inherently limited and biased,
we discussed some existing GDLs, desirable properties of GDLs for AGI
testing, and how existing competitions could be turned into more general competitions.

We hope that this paper can spark and sustain interest into addressing the general AI problem directly within the Game
AI and Computational Intelligence and Games communities, and in developing new challenging
game-based AI competitions.


 




\section*{Acknowledgments}
This work was funded in part by SNF grant number 200020-122124/1.

\bibliography{agigames}

\end{document}